# A Primer on Word Embeddings: AI Techniques for Text Analysis in Social Work

Brian E. Perron[1], Kelley A. Rivenburgh[1], Bryan G. Victor[2], Zia Qi[1], Hui Luan[3]

[1]School of Social Work, University of Michigan

[2]School of Social Work, Wayne State University

[3]School of Social Development, Department of Social Work, Tianjin University of Technology

## Author Note

Brian E. Perron https://orcid.org/0009-0008-4865-451X

Bryan G. Victor https://orcid.org/0000-0002-2092-912X

Hui Luan https://orcid.org/0000-0002-7033-9240

**Abstract**

Word embeddings represent a transformative technology for analyzing text data in social work research, offering sophisticated tools for understanding case notes, policy documents, research literature, and other text-based materials. This methodological paper introduces word embeddings to social work researchers, explaining how these mathematical representations capture meaning and relationships in text data more effectively than traditional keyword-based approaches. We discuss fundamental concepts, technical foundations, and practical applications, including semantic search, clustering, and retrieval augmented generation. The paper demonstrates how embeddings can enhance research workflows through concrete examples from social work practice, such as analyzing case notes for housing instability patterns and comparing social work licensing examinations across languages. While highlighting the potential of embeddings for advancing social work research, we acknowledge limitations including information loss, training data constraints, and potential biases. We conclude that successfully implementing embedding technologies in social work requires developing domain-specific models, creating accessible tools, and establishing best practices aligned with social work's ethical principles. This integration can enhance our ability to analyze complex patterns in text data while supporting more effective services and interventions.

**A Primer on Word Embeddings: AI Techniques for Text Analysis in Social Work**

Social work researchers regularly encounter vast collections of text data, including case notes, interview transcripts, policy documents, social media posts, and open-ended survey responses. Traditionally, analyzing these materials has relied heavily on manual coding and basic keyword searches, which are time-intensive and limited in capturing the subtle meanings and contexts within the text. However, recent advances in natural language processing (NLP) - the branch of artificial intelligence focused on helping computers understand human language - have introduced powerful tools that open up new opportunities for analyzing text data in social work research. Among these tools, word embeddings stand out as a fundamental technology that can dramatically enhance our ability to analyze and understand text-based information.

Text embeddings are a method for representing language in a way that helps computers analyze and interpret relationships between words and concepts. In social work research, this technique can be used to identify patterns in case notes, interview transcripts, and other written materials. By organizing related concepts closer together in a conceptual space, text embeddings enable more meaningful analysis of contextual relationships, moving beyond simple keyword searches. This approach can support the exploration of themes, connections, and trends in social work data.

The recent surge in artificial intelligence, particularly the emergence of large language models (LLMs) like ChatGPT and Claude, has brought renewed attention to text embeddings and created confusion about their role and purpose. While embeddings are indeed a fundamental technology used within LLMs, they are distinct from the capabilities that allow generative AI models to engage in conversation or create new text. Instead, embeddings are a specialized tool for measuring and comparing the similarity between pieces of text. In contrast, LLMs are systems that *generate* new content. This is a critical distinction because embeddings, on their own, offer unique and valuable capabilities for social work research, particularly in analyzing and understanding patterns in existing unstructured text data, without the need for a complex AI system.

Embeddings are also powerful tools for interacting with other AI systems for building various types of applications. For example, a common usage is used with a system called retrieval augmented generation, which is akin to *"chat with your docs."* This means you can use embeddings in conjunction with an LLM to ask a collection of documents for specific information using natural language queries (see Perron, Hiltz, Khang, & Savas, 2024).

The use of word embeddings in social work research is vast and promising. For instance, researchers analyzing case notes could use word embeddings to automatically identify patterns in client experiences that might not be immediately apparent through manual review. When analyzing policy documents, word embeddings can help researchers track how concepts evolve across different versions or jurisdictions. When organizing scientific literature, word embeddings can facilitate systematic reviews by automatically clustering related papers and identifying thematic relationships across different studies (e.g., Luan, Perron, Victor, Wan, Niu, & Xiao, 2024), helping researchers map the intellectual landscape of social work research more efficiently. In community-based research, these tools can help process and analyze large volumes of social media posts or community feedback to identify emerging social issues or community needs. Moreover, word embeddings are often critical in building domain-specific chatbots (see Perron et al. 2024a) and other AI-based support tools.

This methodological paper provides a comprehensive introduction to word embeddings for social work researchers. We begin by explaining the technical foundations of word embeddings, including how they convert text to numbers and measure semantic similarity. We then explore different types of embedding models and their applications, discussing how to prepare text data and visualize embedding relationships. The paper concludes by examining practical applications in research and administrative workflows, addressing limitations, and suggesting future directions for embedding technologies in social work. Throughout, we provide concrete examples from social work practice and research to illustrate key concepts and demonstrate practical implementation strategies.

**Technical Foundations of Word Embeddings**

Before exploring how word embeddings can enhance social work research, it's essential to understand their fundamental mechanics. At their core, word embeddings are a mathematical technique for representing text in a way that captures meaning and relationships. This section breaks down the technical concepts into accessible components, starting with the key challenge of converting text into numbers that computers can process.

For AI systems to analyze text, we must first bridge the gap between human language and computer processing. Computers operate on numbers, not words, so our first task is converting text into numerical form. We'll examine two approaches to this conversion: rule-based methods and term-frequencies.

**Converting text to numbers**

Before computers can analyze text data, we need to convert written documents into numbers that computers can process. This fundamental challenge is particularly relevant in social work research, where vast amounts of clinical documentation exist in text form. Consider research involving the unstructured text in child welfare records. While state child welfare agencies have large collections of structured administrative data, these records often lack the nuance needed to understand problems documented in case notes and reports. The volume of records makes manual review impractical at scale. Two approaches have addressed this problem: rule-based dictionary methods and term-frequencies.

The first approach uses a specialized dictionary of terms to flag relevant documents. For example, Sokol and colleagues (Sokol, Victor, Piellusch, Nielsen, Ryan, & Perron, 2020) created a comprehensive dictionary of firearm-related terms (e.g., "gun," "firearm," "rifle") to identify documents mentioning firearms in child welfare investigation summaries. Perron, Victor, Ryan, Piellusch, & Sokol (2022) also used this approach for identifying opioid mentions – a strategy that falls under a broader methodological category that is referred to as clinical or qualitative text mining (Cohen & Hersh, 2005; Henry, Carnochan, & Austin, 2014) -- that is, using the dictionary as a filter to identify relevant documents within large collections of records efficiently. These documents are then manually reviewed to confirm they meet the inclusion criteria for the study, vastly reducing the amount of data requiring manual review.

The rule-based method can also be part of a statistical model. For instance, rather than just flagging documents for review, Perron et al. (2019) and Victor et al. (2021) used dictionary terms as variables in statistical models to automatically classify child welfare records containing substance-related problems and domestic violence, respectively.

The second approach analyzes how frequently words appear across documents without making prior assumptions about important terms. This involves constructing a TF-IDF (Term Frequency-Inverse Document Frequency) matrix. Think of TF-IDF as a way to measure how important a word is to a document in a collection. Words that appear frequently in a specific document but rarely in other documents get higher scores, helping identify distinctive terminology for each type of case. Both Perron et al. (2019) and Victor et al. (2021) used this method alongside their dictionary approaches to compare performance. The term-frequency method can identify relevant patterns that expert-created dictionaries might miss, as it lets the patterns emerge naturally from the data. This approach showed to be more effective than rule-based methods, with results that approximated human reviews. Yet, both approaches are limited because they ignore the context of a given word or phrase.

Word embeddings represent a significant advancement in how we convert text to numbers, moving beyond simple counting to capture deeper relationships in language. This approach creates a sophisticated mathematical representation of words and documents that preserves meaning and context - something particularly valuable for analyzing social work documentation.

Think of word embeddings like creating a detailed map of concepts. Just as a city map shows how different neighborhoods relate to each other spatially, word embeddings show how different terms in social work practice relate to one another mathematically. Terms frequently used in similar contexts appear closer together in this mathematical space. For example, in social work documentation, terms like "housing assistance," "emergency shelter," and "transitional housing" cluster near each other because they often appear in similar contexts when discussing housing services. Similarly, terms related to mental health services, such as "therapeutic support" and "counseling," form their own cluster, reflecting their distinct but related role in service provision.

Embeddings work at multiple levels - from individual words to entire documents. At the word level, related concepts like "anxiety" and "depression" appear close together because they're used in similar clinical contexts, while unrelated terms like "anxiety" and "umbrella" would be far apart. At the document level, two case notes describing similar client situations would be mathematically closer than notes about different types of cases. This ability to capture meaningful relationships represents a significant advance over earlier methods that simply counted word occurrences or checked against predefined dictionaries.

Embeddings are not used by themselves in research but are part of what is referred to as "downstream tasks" - specific applications that build on the mathematical representations created by embeddings. For example, Perron et al. (2024a) used embeddings to construct a question-answering system about academic policies in a school of social work. Luan et al. (2024) used embeddings to conduct a topical analysis and classify a large collection of scientific abstracts on the topic of left-behind children. The utility of embeddings is further demonstrated by Semantic Scholar, the largest open repository of scientific metadata. This platform not only uses embeddings to provide paper recommendations but also includes them as standard metadata alongside traditional fields like author names and publication dates, providing new ways to effectively organize and access scholarly literature. Before exploring these applications in detail, understanding these core principles of embeddings provides essential context for what follows.

**Core Principles of Word Embeddings**

Word embedding models can transform words, sentences, and even entire documents into *vectors*, which are lists of numbers that locate each word in a complex mathematical space. Unlike rule-based or dictionary methods that rely on predefined lists of words and their meanings, word embeddings capture relationships between words based on their usage in large bodies of text. These vectors typically use between 100 and 1,500 numbers to represent each word. For example, the word "community" might be represented as a vector like [0.12, -0.45, 0.78, ..., 0.34], where each number corresponds to a specific dimension of the semantic space. These dimensions reflect different linguistic or contextual features of the word, allowing the vector to capture nuanced relationships. Words that are contextually or semantically related, such as "neighborhood" and "support," will have vectors with similar values, positioning them closer together in the semantic space.

To illustrate how word embeddings can enhance social work research, consider the work by Sokol et al. (2020) on identifying firearm-related content in child welfare records. In their study, they used a specialized dictionary of firearm-related terms—such as "gun," "firearm," "rifle"—to flag documents mentioning firearms in child welfare investigation summaries. While this rule-based method effectively filtered documents for manual review, it had limitations. The approach relied heavily on predefined terms and could miss documents that discussed firearms using alternative language or colloquial expressions not included in the dictionary. Additionally, it couldn't capture the context in which firearms were mentioned, such as whether a firearm was present in the home or involved in an incident.

The contextual similarity is important when dealing with words that have multiple meanings or homonyms. For instance, names like "Remington," "Colt," and "Winchester" can refer to firearm brands or people's names. In traditional keyword searches, a document mentioning "Colt" could be mistakenly flagged as firearm-related even if it's actually about a person named Colt. However, word embeddings analyze the context in which "Colt" appears. If "Colt" is used alongside terms like "school," "attendance," and "behavior," it would be understood as a person's name and placed in a different region of the semantic space. Conversely, if "Colt" appears near words like "ammunition," "shooting," or "weapon," it would be associated with firearms.

We can use different algorithms to measure similarity between texts and to find "nearest neighbors." This means a document that might use alternative language, slang, or has ambiguous terms, the embedding model can capture the contextual similarity and cluster those with related documents. This allows us to capture a broader range of relevant documents, reducing the risk of missing critical information due to vocabulary variations or ambiguous terms. Clustering documents in semantic space also allows us to discover patterns and relationships that might not be immediately apparent through manual review or simple keyword searches. For example, we might find that documents discussing firearms also frequently mention factors like domestic violence, substance abuse, or mental health issues. By visualizing these clusters and their proximities, we can gain deeper insights into the co-occurrence of risk factors, which is crucial for developing comprehensive intervention strategies.

In sum, word embeddings enable us to move beyond surface-level text analysis to uncover richer, contextual relationships embedded in language. This capability is particularly valuable across various fields where understanding nuanced meanings and patterns is essential for effective communication and decision-making. By leveraging word embeddings, we can more effectively analyze large volumes of unstructured text data, discover hidden patterns, and ultimately enhance the quality and impact of our research and applications. This opens up opportunities for more sophisticated natural language processing tasks, such as semantic search, sentiment analysis, and topic modeling, thereby advancing our ability to interpret and utilize textual information in meaningful ways.

## Types of Word Embedding Models

The development of word embeddings has advanced rapidly in recent years. Early models, such as Word2Vec introduced around 2013 (Mikolov, Sutskever, Chen, Corrado, & Dean, 2013), laid the foundation by representing words as numerical vectors based on the words they frequently appear alongside in large collections of text. This approach captured basic relationships between words, allowing us to see that words like "king" and "queen" are related because they often occur in similar contexts. However, these early models had limitations—they treated each word the same regardless of the specific context in which it appeared, lacking the ability to account for the nuances of language use in different situations.

Contemporary embedding models have evolved to address these limitations by incorporating the context surrounding each word. For example, OpenAI's Ada embedding model is trained on vast amounts of text from books, articles, and websites (OpenAI, 2024). During training, the model learns to predict missing words in sentences. For instance, if given the sentence "The counselor advised the client to consider ____ strategies," the model learns to predict words like "coping" or "intervention" based on the context provided by the surrounding words. By repeatedly performing this task on millions of sentences, the model learns not only the meanings of individual words but also how context influences those meanings. These modern models use an approach called "encoder architecture." In simple terms, an encoder reads text and converts it into a numerical form that captures both the meaning of the words and the context in which they appear. This process allows the model to create embeddings for not just individual words, but also for sentences and entire documents, reflecting the nuances and complexities of language more accurately.

By taking context into account, these contemporary embedding models produce more precise and meaningful representations of text. This advancement opens up a broader range of applications. For example, they enable more effective semantic search, where a system can find relevant information based on the meaning of a query rather than just keyword matching. They improve text classification tasks by accurately capturing the sentiment or topic of a document, even when subtle language is used. This is particularly valuable when analyzing large volumes of unstructured text data, such as case notes or interview transcripts, where understanding the nuanced meaning is crucial.

Word embedding models can fall under two general categories: general purpose models and domain-specific models. These are described in turn.

**General purpose models and their limitations for social work research**

General purpose embedding models are trained on vast collections of text from diverse sources—including books, websites, academic papers, and news articles—allowing them to understand language patterns across many fields and contexts. During this initial training, the model learns how different words and concepts relate to each other, creating a map of meanings that it can use later when looking at new documents. Then, when a user gives the trained model new documents to convert into embeddings, the model uses what it learned during training to convert these documents into a numeric format that captures their meaning based on patterns learned during training. Say for example that during its initial training on millions of documents, the model learned that 'anxiety' and 'depression' often appear in similar contexts and are closely related concepts. When the model later processes new documents and encounters these terms, it will place them close together in its numeric representation of the text, making it easier to identify documents discussing related mental health topics in later analysis.

OpenAI's text-embedding models (OpenAI, 2024) have emerged as a de facto standard among these models due to their strong performance and ease of use. These models capture broad language understanding and can effectively analyze texts across disciplines, making them particularly valuable for analyzing public policy documents, published research papers, and government reports. For example, when analyzing housing policy documents, it can recognize that phrases like "residential instability" and "housing market displacement" describe related housing security issues, even when they use different terminology. When users provide it with new documents, the pre-trained model can convert those documents into a numeric representation in a way that can help researchers identify similar policy approaches across different jurisdictions or track how policy language evolves, such as connecting traditional terms like "low-income housing" with newer concepts like "housing affordability" or "inclusive development." This capability is especially valuable for systematic reviews of policy literature or when comparing policy frameworks across different regions or periods.

While general-purpose embedding models offer powerful capabilities, they come with two significant limitations for social work research. First and most critically, models like OpenAI's embeddings require sending text to external servers for processing, making them unsuitable for analyzing confidential client information, protected health data, or any sensitive materials that cannot leave secure institutional environments (Perron, Luan, Victor, Hiltz-Perron, & Ryan, 2024). This restriction means researchers working with case notes, clinical assessments, or other private documentation need alternative solutions that can be deployed within their secure systems.

Second, despite their broad knowledge, these models may not fully grasp the nuanced terminology and specialized concepts central to social work practice. While they may understand terms like "assessment" or "intervention" generally, they often miss their specific professional implications. For instance, they might not recognize that "strengths-based assessment" represents a distinct philosophical approach or that "trauma-informed care" encompasses specific principles beyond general trauma awareness. These limitations become particularly apparent when analyzing professional documentation.

**Domain-specific models**

Domain-specific models are embedding models trained or fine-tuned on specialized professional documentation to understand field-specific terminology and contexts better. We are unaware of embedding models specifically fine-tuned for social work. However, several promising models have been trained on related healthcare and clinical documentation. These models are primarily available through Hugging Face. This open-source platform is a central hub for sharing and accessing machine learning models. In other words, it is a repository of open-source models. Researchers can access general-purpose and domain-specific models through HuggingFace's model repository. Domain-specific models include clinical-longformer (Li, Wehbe, Ahmad, Wang & Luo, 2023) and Pubmedbert (Gu et al., 2020), trained on extensive collections of medical notes and healthcare research papers.

These models offer two critical advantages over general-purpose embeddings like OpenAI's. First, they can be downloaded and run locally within secure institutional environments, addressing the privacy concerns associated with cloud-based services. Second, they demonstrate a better understanding of clinical terminology, mental health concepts, and healthcare delivery frameworks that overlap with social work practice. For instance, these models better understand that "CBT" refers to "cognitive behavioral therapy" or that "ACEs" means "adverse childhood experiences" rather than other possible meanings. While not perfect substitutes for social work-specific models, they offer advantages over general-purpose embeddings when analyzing literature and documentation that intersects with healthcare and clinical practice. This points to a significant opportunity for future research: developing embedding models specifically trained on social work documentation to capture the field's unique theoretical frameworks, practice approaches, and professional terminology.

**Choosing Embedding Models**

Selecting an appropriate embedding model for social work research requires careful consideration of several key factors. Many high-quality embedding models are available today, each offering different trade-offs that should align with your research requirements and constraints. A primary consideration is data privacy and security. Some models require sending text to external servers for processing, while others can be run entirely within your institutional infrastructure. This distinction becomes crucial when working with confidential client information, protected health information, or sensitive research data. Cost structure represents another critical factor, as models vary between pay-per-use services and free, open-source options. However, remember that "free" models often require more substantial computational resources and technical expertise. Other factors to consider include:

- Data privacy and security requirements
- Available budget and cost constraints
- Technical infrastructure and expertise
- Processing speed requirements
- The scale of text analysis needed

The decision process should begin by examining these criteria within your research context. For instance, a project analyzing public policy documents might prioritize processing speed and ease of implementation, while research involving client records would prioritize data privacy and security. This systematic evaluation ensures the selected model meets technical requirements and aligns with social work research's ethical and practical constraints.

## Measuring Embedding Similarity in Practice

Once text is converted into embeddings, researchers need effective methods to measure similarity between these mathematical representations. Cosine similarity is the most common approach, measuring how closely aligned two embeddings are in the mathematical space. This method produces values between -1 and 1, where:

- 1.0 indicates perfect similarity (identical meaning)
- 0.0 indicates no relationship
- -1.0 indicates opposite meaning (rarely seen in practice)

To demonstrate this approach, we use the BAAI/bge-m3 embedding model (Chen et al., 2024), which has gained significant adoption with over 3 million downloads in a single month at the time of writing. Consider a practical example: identifying case notes containing references to housing instability. We can begin by establishing a base sentence as our comparison point: "Client requires emergency housing placement due to domestic violence situation." Assume the following case note snippets from other cases, along with their similarity scores:

> (0.804) "Client seeking temporary housing and support services while fleeing unsafe living conditions."
>
> (0.794) "Client needs immediate shelter assistance after being expelled from current residence due to safety concerns."
>
> (0.718) "Client requests assistance with basic needs including food, utilities, and stable accommodation."
>
> (0.625) "Client working with employment specialist to improve job search skills and interview techniques."
>
> (0.575) "Client demonstrates good progress in group therapy sessions focused on stress management."

The similarity scores reveal clear patterns in clinical documentation. The highest similarity scores (0.804 and 0.794) appear when notes describe urgent housing needs combined with safety concerns, even with slightly different phrasing. Both top-scoring notes share key elements with our reference: immediate need, housing instability, and safety risks. A strong similarity (0.718) persists when notes discuss housing within a broader context of basic needs, though the urgency element is reduced. The score decreases notably (0.625) when notes shift to employment services, though it maintains some similarity due to the shared focus on concrete service provision. The lowest score (0.575) occurs with therapeutic progress notes, which share the least content overlap with housing crises. However, even this lowest score maintains moderate similarity because it reflects common elements of social work practice documentation - client engagement and progress monitoring - which share professional language patterns even when discussing different service domains. This pattern of decreasing similarity scores accurately reflects how the model distinguishes between crisis-level housing needs and other social work interventions while still recognizing their shared professional context.

Consider another example in a different sector of social work, elder care. Practitioners may need to identify common patterns in clients experiencing medication management challenges in this context. The following source sentence was used for comparison: "Senior client demonstrates increasing difficulty with medication management and has missed several important doses." Suppose the following are excerpts from electronic health records coupled with their similarity scores relative to the source sentence.

(0.864) "Client demonstrates persistent difficulty managing medications, missing multiple doses, and shows increasing confusion with prescription schedules."

(0.670) "Client needs an updated transportation schedule for medical appointments."

(0.646) "Client's kitchen contains multiple expired food items, and checkbook shows evidence of mathematical errors in recent transactions."

(0.611) "A report was made about suspected elder abuse."

(0.539) "Patient is attending art appreciation classes at a local community college."

Analysis of these similarity scores reveals patterns in the records. The highest score (0.864) reflects direct context overlap in medication adherence and shared key terms ("demonstrates," "difficulty," "medications," "doses"), with additional detail about confusion with prescriptions. The second highest score (0.670) appears with transportation scheduling, likely elevated due to its connection to medical care management and appointment adherence. The moderate score (0.646) indicates alignment in themes of declining self-management capacity, though in different domains (food safety and financial management). A similar score (0.611) emerges with elder abuse reporting, sharing the serious concern for elder wellbeing though in a different context. The lowest score (0.539) retains only the elderly demographic context while focusing on recreational programming.

**Interpreting Similarity Scores**

Similarity scores measure the degree of alignment between textual data, enabling researchers to unique opportunities to explore patterns, relationships, and classifications. Two main approaches are commonly used to interpret these scores: applying predefined thresholds or adopting a relative comparison strategy. These methods often draw on frameworks such as Cohen's thresholds for effect sizes, adapted to suit the unique requirements of textual analysis.

Cohen's (1992) thresholds for effect sizes provide a useful starting point. Originally developed for standardized differences, such as Cohen's *d*, these thresholds categorize effects as small (0.2), medium (0.5), and large (0.8). While these thresholds were developed for specific effect size measures, they are commonly adapted as a general reference point in various contexts. In the case of similarity scores, these benchmarks can have a parallel interpretation – that is, as small effects for values below 0.2, medium effects for values between 0.2 and 0.5, and large effects for values above 0.5, acknowledging that this is an extension of Cohen's original framework.

Researchers can set specific thresholds to identify strongly related concepts or descriptions. For example, when studying how trauma-informed practices are described across different settings, a very high similarity score (above 0.80) might indicate core principles being consistently expressed, even when using different terminology. Moderate scores (0.60-0.80) often emerge when comparing related but distinct aspects of practice, such as different components of family support interventions or various approaches to community engagement. While this threshold approach offers clear decision rules, it may miss meaningful relationships that fall just below arbitrary cutoff points.

Alternatively, researchers might examine relative patterns of similarity using a relative, or *best fit,* approach, building on Luan et al.'s (2024) application in scientometric analysis. Luan et al. demonstrated how word embeddings can be used to calculate cosine similarities between texts and predefined categories, selecting the category with the highest similarity score as the best match. This method acknowledges that real-world social work phenomena rarely fit perfectly into predefined categories but can still be meaningfully classified based on relative similarity. For instance, when analyzing client narratives, a description might show moderate similarity scores across multiple service categories, but Luan et al.'s approach would identify the most closely aligned category based on the highest similarity score. While this relative approach is more flexible than strict thresholds, it requires careful consideration of the predefined categories and may sometimes force classification into categories that do not have a strong conceptual fit.

These computational methods offer helpful ways to analyze large amounts of text data alongside traditional research approaches. However, it's important to understand what these tools can and cannot do. Similarity scores and word embeddings are mathematical ways to compare text - they can identify patterns and relationships that might be hard to spot manually, but they don't truly understand meaning the way humans do. Their usefulness depends on three key factors: how well the embedding model captures language nuances, whether the categories we are comparing against make sense for our research questions, and how well we connect the mathematical results to real-world knowledge and expertise. These tools work best when used to support and enhance human analysis rather than replace it.

**Word embeddings at scale**

When conducting text similarity analysis, the scale of the document collection influences the tools required. For small-scale analyses—such as comparing a few hundred or a few thousand text documents—a researcher can directly calculate similarity scores between embeddings using standard software tools like Python libraries (Python, 2023).

However, as the collection grows to include tens or hundreds of thousands of documents, or even millions, more sophisticated tools become necessary to handle the increased computational demands efficiently. In large-scale similarity searches, researchers typically rely on specialized vector databases and similarity search libraries. Vector databases like Pinecone, Weaviate, or Milvus are designed to effectively store and manage large numbers of embeddings. They are optimized for vector data, ensuring that operations like inserting, updating, or retrieving embeddings are performed efficiently.

To compute similarities and retrieve relevant documents from large datasets, libraries like FAISS (Facebook AI Similarity Search) are used (Douze et al., 2024). FAISS offers advanced algorithms for rapid similarity computation and searching within extensive collections of embeddings. FAISS enables fast and accurate identification of similar texts, optimizing the process for the high-dimensional data produced by embedding models. For example, Kaushal et al. (2023) used FAISS to analyze underlying patterns in over 46,000 social media posts about COVID-19 vaccines. By creating a FAISS index of their entire dataset's text features, they were able to complete a detailed exploration of thematic clusters and sentiment patterns that would have been too computationally intensive through traditional methods.

While tools like FAISS enable swift retrieval of similar documents based on measures like cosine similarity, they may sometimes miss more nuanced matches due to limitations in capturing deeper contextual relationships. This is where a technique called *reranking* comes into play (Google Cloud, 2024). Reranking involves taking an initial set of candidate documents—retrieved using basic similarity measures—and reordering them based on a more sophisticated analysis to improve relevance. For example, after initially retrieving the top N most similar documents, you apply a more advanced model or set of criteria to these candidates, such as deeper linguistic analysis or leveraging a more powerful language model. This reassessment prioritizes documents that are more contextually relevant, capturing subtle relationships and meanings that basic similarity measures might overlook.

## Text Preparation Principles

Text preparation for embeddings marks a fundamental shift from traditional text analysis approaches. Traditional approaches to data preparation prior to the use of AI-based methods like machine learning involve modifying text in different ways, such as removing stop words - frequently occurring words like "the," "is," "are," "by," "with," and "without" that serve grammatical functions but were historically considered to lack meaningful content. For example, a social worker's case note stating "Client is currently living with family members due to housing instability" would traditionally be reduced to isolated key terms: "client," "living," "family," "housing," "instability." This approach would lose crucial meaning - the distinction between "living with family" versus "living without family" would be lost entirely since connector words would be removed. These subtle but vital differences in meaning demonstrate why traditional stop word removal can eliminate essential context from human communication.

Embedding analysis takes a dramatically different approach. Instead of stripping away common words and treating each term in isolation, it preserves natural language patterns that create meaning - including critical stopwords. This preservation is crucial because embedding models don't simply catalog words but represent the relationships inherent in how language is naturally used. Consider these contrasting phrases in social work documentation:

- "Client living with family" (indicating support system)
- "Client living without family" (indicating isolation)
- "Client working with counselor" (indicating engagement)
- "Client working without counselor" (indicating potential disengagement)

These phrases have dramatically different meanings in social work practice, distinctions that would be lost if we removed stopwords like "with" and "without." Embedding models preserve these crucial meaning-making words and their relationships.

While they retain important connector words that help to preserve meaning, the use of word embedding models often requires the researcher to "split" or "chunk" their text documents into smaller units. This process of splitting or chunking the text is necessary because most embedding models have a maximum length of text they can process at once, typically measured in tokens (roughly equivalent to word parts). When working with long documents, splitting them into smaller, meaningful chunks ensures that each piece stays within these length limits while preserving coherent ideas and context. Additionally, having smaller text segments allows for more precise information retrieval later on, since you can pinpoint relevant sections of a document rather than having to work with entire documents at once.

Modern tools offer several sophisticated approaches to creating these meaningful chunks. Available through libraries like LangChain (Chase & LangChain Contributors, 2024) and LlamaIndex (Liu & LlamaIndex Contributors, 2024), text splitters can automatically segment documents while preserving context. These tools use recursive character splitting or sliding windows to keep related information together. For example, when processing a lengthy case note, a text splitter might maintain the connection between a client's housing situation and associated mental health symptoms by creating overlapping chunks.

Alternatively, LLMs can assist in performing intelligent summarization or extraction. Instead of merely dividing text (see Luan et al., 2024). As examples, an LLM can:

1. Extract structured information: "Primary Need: Housing instability; Current Status: Temporary family housing; Risk Factors: Job loss, anxiety."

2. Create focused summaries: "Client experiencing housing instability after a job loss, currently in temporary family housing arrangement causing anxiety."

3. Generate standardized chunks: Converting documentation styles into consistent formats while preserving critical information.

Each approach serves different analytical needs while maintaining the crucial context for the embedding models. The choice between these methods often depends on your research goals and the nature of your documentation. The distinction between signal and noise in social work documentation remains crucial regardless of the chunking method. Whether using automated text splitters or LLM-based approaches, we want to preserve meaningful content while removing actual noise, such as standardized headers, timestamps, routing information, and boilerplate text that appears in every document. Noise can be any bit of text that does not contribute to the intended meaning of the chunk. For example, Luan et al. (2024) used an embedding approach to classify scientific articles into pre-defined topics. They reported summarizing abstracts to remove all text that was unrelated to the actual subject of the research.

A common mistake is attempting to maximize the information in each embedding by cramming as much text as possible into each chunk. This approach misunderstands how embedding models work – that is, the models don't store every word in isolation but rather create a mathematical representation of meaning in a high-dimensional space. A focused, coherent chunk often produces more useful embeddings than a longer, unfocused one. The goal of text preparation, regardless of the method chosen, is to create chunks that:

- Represent complete, coherent thoughts or assessments;

- Maintain natural language patterns and context;
- Remove true noise while preserving meaningful content;
- Support comparison and analysis across documents.

**Visualizing Word Embeddings**

Understanding and communicating patterns in word embeddings presents a unique challenge for social work researchers. As discussed earlier, embedding models typically represent text using vectors with hundreds or even thousands of dimensions - far beyond what humans can directly visualize or interpret. For instance, a popular embedding model like BAAI/bge-m3 represents each piece of text with 1,024 separate numerical values, while OpenAI's text-embedding-3-small uses 1,536 dimensions. To make these complex high-dimensional relationships interpretable, researchers typically turn to t-SNE (t-Distributed Stochastic Neighbor Embedding), which has emerged as the standard method for visualizing word embeddings (van der Maaten & Hinton, 2008).

t-SNE works by converting high-dimensional distances between points into probabilities that represent similarities. The algorithm then creates a lower-dimensional representation (typically 2D or 3D) that preserves these similarity relationships as much as possible. Unlike simpler dimensionality reduction techniques like principal components analysis, t-SNE is particularly good at preserving local structure in the data, making it well-suited for visualizing the nuanced relationships typically found in word embeddings. For instance, when analyzing case notes, t-SNE might reveal clusters of documentation related to different types of interventions or client needs. Unlike linear methods, t-SNE can capture complex, non-linear relationships in the data. This is particularly important for word embeddings, where semantic relationships rarely follow simple linear patterns. t-SNE also tends to create clear visual separation between different groups of points, making it easier to identify potential patterns in the data. However, researchers must interpret these clusters cautiously, as t-SNE can sometimes exaggerate the separation between points.

We analyzed similarities between social work licensing examinations from the United States and China to demonstrate a multilingual example. Our analysis included 50 questions from the Association of Social Work Board (ASWB) master's level examination preparation materials (see Victor, McNally, Qi, & Perron, 2024; Victor, Kubiak, Angell, & Perron, 2023; Association of Social Work Boards, 2024) and 80 questions from the Chinese National Social Work Association's applied knowledge test (National Social Worker Professional Exam Question Compilation Group, 2024). Using only the question stems without answer choices, we used Jina AI's state-of-the-art multilingual embedding model, selected for its robust performance with both English and Chinese texts.

We applied t-SNE analysis to reduce the high-dimensional embeddings to two components to visualize the relationships between these questions. The resulting scatterplot revealed distinct clustering between Chinese and English examination questions, suggesting fundamental differences in how these exams approach social work knowledge assessment.

**[INSERT FIGURE 1 ABOUT HERE]**

Within each language cluster, the visualization revealed meaningful patterns in how questions were grouped based on their content. Though every exam question was unique, those addressing similar topics appeared closer together in vector space. An inspection of the patterning helps build confidence in the performance of the embedding model regarding the expected patterning of topics. For instance, in the ASWB exam cluster, questions that appeared close together shared thematic elements:

- Questions 33 and 38 are close together in semantic space, and both questions are about policies and procedures involving the handling of client records.
- Questions 1 and 29 are distant from each other. The former addresses a community planning issue, whereas the latter pertains to individual-level spiritual assessment.

The Chinese-language cluster showed similar content-based groupings:

- Questions 17 and 23 appeared close together, both addressing employment-related issues
- Question 48, positioned separately from 17 and 23, focuses on organizational budgeting methods

These patterns demonstrate how semantic similarity in question content translates to proximity in vector space, regardless of the language of origin. This clustering helps us understand the topical organization of social work knowledge assessment in both contexts.

# Word Embeddings in Research and Administrative Workflows

As emphasized at the outset of this article, word embeddings are a way to represent data numerically. We can use these embeddings within larger research and administrative workflows by embedding unstructured text data. This section provides examples of how embeddings are applied to different data types for a specific purpose. For each use case, we describe a particular problem, our data source, and embeddings' role in addressing those data problems.

**Semantic Search**

Traditional keyword-based search methods often miss relevant results because they lack the ability to understand context or recognize related concepts. This limitation is especially challenging when identifying subject matter experts or relevant research based on nuanced descriptions rather than exact keywords. For example, selecting peer reviewers for specialized research articles or external evaluators for tenure and promotion cases traditionally involves manually reviewing faculty profiles and CVs across institutions. This time-intensive task risks overlooking qualified candidates whose expertise may be described in varied terms.

To address this, we developed a proof-of-concept application for search retrieval from a curated database of over 1,000 faculty biographies from the top 20 schools of social work as determined by U.S. News and World Report. Each biography was standardized using an LLM, and embeddings were computed using Jina AI embeddings (v3; Jina AI, 2024). Jina AI is a general-purpose, open-source embedding model trained on extensive public datasets to capture broad linguistic patterns, making it effective for identifying semantically similar content across various applications.

Users enter a research topic as a search query, which is then converted to an embedding and compared with the embedded faculty biographies. Cosine similarity is used to identify biographies with the highest similarity scores. The following screenshot (see Figure 2) shows the tool with a search query, "Cancer treatment and support for young adults," displaying the top matches.

**[INSERT FIGURE 2 ABOUT HERE]**

The top result, Anao Zhang, described a "primary research interest in psycho-oncology and adolescent and young adult cancer survivorship," achieving a similarity score of 64.23%. While this record does not contain the exact language of the query, the embedding model accurately identified it through semantic similarity (see Figure 3).

[INSERT FIGURE 3 ABOUT HERE]

The second-ranked individual, Bradley J. Zebrack, mentions experience in "both pediatric and adult oncology" and involvement in "peer support programs for adolescent and young adult cancer survivors," with a similarity score of 57.99%. A third biography, by Qi Chen, describes "a research project funded by the American Cancer Society examining social media use and mental health among young adult cancer survivors," yielding a similarity score of 53.36%.

This semantic understanding dramatically improves the efficiency and effectiveness of expert identification, ensuring that relevant expertise isn't overlooked due to variations in terminology. The system can recognize conceptually related work even when described using different academic or professional language, making it particularly valuable for interdisciplinary research where terminology often varies across fields.

**Clustering and Topic Modeling**

Word embeddings have emerged as a powerful foundation for clustering and topic modeling of unstructured text, offering significant advantages over traditional approaches. While earlier methods like Latent Dirichlet Allocation, Term Frequency-Inverse Document Frequency, Non-negative Matrix Factorization , and probabilistic Latent Semantic Analysis have been used for clustering and topic modeling, embedding-based approaches provide a more intuitive and robust framework for understanding textual relationships. As previously described, traditional text analysis methods rely on word frequency counts or simple presence/absence indicators, failing to capture the rich semantic relationships between concepts. For example, TF-IDF might identify that terms like "housing" and "shelter" appear frequently in a document. Still, it cannot understand that these terms are semantically related or that "temporary accommodation" refers to a similar concept. In contrast, embedding-based approaches preserve these

24crucial semantic relationships, allowing for a more nuanced analysis of how ideas and themes relate to text data.

Recent research demonstrates the practical value of embedding-based clustering in social work research. For instance, Luan et al. (2024) applied this approach to analyze research literature on left-behind children. Their process began by creating embeddings of article abstract summaries, which captured each paper's key concepts and relationships. These embeddings were then used to perform a cluster analysis, a statistical method that groups similar items while ensuring items in different groups are as distinct as possible. Cluster analysis acts like an automated sorting system, examining the mathematical relationships between embeddings to identify natural groupings in the data.

In the research literature, these clusters often represent distinct research themes or approaches. For example, in the Luan et al. (2024) study, the cluster analysis suggested distinct papers focusing on different aspects of left-behind children's experiences – some clusters centered on educational outcomes, others on mental health impacts, and laws and policies regarding guardianship. Each cluster represented a collection of papers that addressed similar underlying concepts and research questions while using potentially different terminology. Its ability to identify these thematic relationships even when papers use different terminology or approaches makes embedding-based cluster analysis particularly powerful. Traditional keyword-based approaches might miss the connection between a paper discussing "academic achievement" and another focusing on "educational attainment," but embedding-based clustering recognizes these conceptual similarities. This capability is especially valuable in social work research, where similar concepts might be described using different terminology across different practice contexts or theoretical frameworks.

**Retrieval Augmented Generation**

Another area where word embeddings can help to advance social work research relates to the increasing use of large language models (LLMs) within disciplinary scholarship (Patton et al., 2023). A common challenge associated with LLM use is their propensity to hallucinate or generate factually incorrect outputs (Victor, Goldkind, & Perron, 2024). Retrieval augmented generation (RAG) is a strategy to correct this problem by using word embeddings to help LLMs access domain-specific knowledge bases. A knowledge base in this context refers to a structured collection of documents, policies, procedures, or other text-based information that contains authoritative content for a particular domain or organization. For instance, a school of social work might create a knowledge base containing its student handbook, course catalogs, field education manuals, and frequently asked questions about academic policies.

RAG first converts this knowledge base into embeddings that numerically represent the semantic meaning of text. When a user poses a question, RAG follows a three-step process:

1. The user's question is converted into an embedding using the same model used to embed the knowledge base

2. The system searches for relevant information by comparing the question's embedding to the pre-computed knowledge base embeddings using similarity measures

3. The retrieved information is then provided to the LLM along with the original question, allowing it to generate a response grounded in accurate, domain-specific knowledge

This process helps correct the fundamental limitation of standard LLMs, which is their tendency to generate plausible but potentially inaccurate responses when dealing with domain-specific information. By connecting the LLM to a curated knowledge base through embeddings, RAG ensures responses are anchored in accurate, authoritative information.

Consider an advising support tool evaluated at a school of social work (Perron et al. 2024a). When a faculty member asks, "How many field hours do students need for their MSW?" this question is converted into an embedding. The system then calculates the similarity between this query embedding and all the pre-computed embeddings from the knowledge base. Because embeddings preserve semantic relationships, chunks of text about field education requirements will have similar vector representations and thus be identified as relevant, even if they use slightly different terminology.

The success of RAG systems depends heavily on how effectively embeddings capture and represent the semantic relationships in the domain-specific content. For instance, in social work education, the embedding model needs to understand that terms like "field placement," "practicum," and "internship" refer to similar concepts. The quality of these embeddings directly impacts the system's ability to retrieve relevant information.

The RAG approach showcases how embeddings can bridge the gap between general-purpose AI models and domain-specific knowledge needs. By using embeddings as the mathematical foundation for information retrieval, RAG systems can provide more accurate and contextually appropriate responses while maintaining the flexibility to handle natural language queries. This makes them particularly valuable for social work applications, where precise and reliable information access is crucial for effective practice and decision-making.

## Limitations and Next Steps

As we discuss the value of embeddings, we must acknowledge their inherent limitations. While embeddings provide necessary tools for text analysis and information retrieval, understanding their limitations is essential for their practical application in social work research and practice. We identify several critical limitations that deserve careful consideration.

**Limitations of Word Embeddings**

A primary limitation of word embeddings is the training data used to create embedding models. Most general-purpose embedding models are trained on broad datasets that may not adequately represent specialized professional terminology or context-specific language patterns. This limitation becomes particularly apparent in social work, where professional terminology and concepts may differ significantly from general usage. Terms like "strength-based approach," "person-in-environment perspective," or "trauma-informed care" carry specific professional meanings that might not be accurately captured by models trained primarily on general text data. This training limitation may also be heightened when working with specialized populations or cultural contexts where embedding models may not adequately capture culturally specific expressions or community-specific terminology.

The static nature of embeddings presents a second limitation. Once an embedding model is trained, its understanding of language relationships remains fixed unless explicitly retrained. This poses challenges in social work, where terminology, best practices, and social issues evolve continuously. New therapeutic approaches, emerging social problems, or shifts in professional terminology need to be accurately represented in existing embedding models, particularly when analyzing contemporary social issues or novel intervention approaches.

Third, embeddings can perpetuate and amplify biases present in their training data. Social work's commitment to social justice and equity makes this limitation particularly concerning. Embedding models trained on historical texts or general social discourse may encode societal biases related to race, gender, socioeconomic status, or other demographic factors. These biases manifest in how the model represents relationships between concepts, potentially influencing downstream applications like information retrieval or analysis. An embedding model might inadvertently reinforce stereotypes about specific communities or create inequitable associations between demographic characteristics and social problems.

Finally, the *black-box* nature of embeddings presents a fifth critical limitation. While we can measure similarities between embeddings, the individual dimensions of these vectors lack clear interpretability. Unlike traditional content analysis, where themes and categories are explicitly defined, humans do not directly interpret the mathematical features captured by embeddings. This lack of transparency can be problematic in social work contexts where clear explanation and accountability are essential, particularly when embeddings are used to support decision-making processes.

Understanding these limitations helps researchers set realistic expectations and develop appropriate strategies for using embeddings alongside other analytical tools. It emphasizes the importance of maintaining access to original text documents alongside their embedding representations, particularly in contexts where preserving full textual detail is crucial. While domain-specific models trained on relevant professional literature can help address some limitations, they require significant resources to develop and maintain. They will still face the fundamental limitation of information loss inherent in the embedding process.

Despite these constraints, embeddings remain valuable tools when used appropriately and with a clear understanding of their limitations. The key is recognizing where embeddings can most effectively support social work practice and research while maintaining awareness of situations where alternative approaches might be more appropriate. This balanced understanding allows us to leverage the strengths of embedding technology while implementing proper safeguards and complementary approaches where needed.

**Advancing the use of embeddings**

The advancement of word embeddings in social work research and practice requires strategic development across several dimensions. While current applications demonstrate promise, broader adoption and effective implementation will require technical skill development and an enhanced understanding of these technologies among social work professionals (see Perron, Victor, Hiltz, & Ryan, 2022).

A key consideration is the accessibility of embedding technologies. Currently, implementations typically require programming expertise, particularly in Python, the dominant language for AI and natural language processing applications. While no-code solutions are emerging that provide user-friendly interfaces—similar to how AMOS provided a graphical alternative to Mplus for structural equation modeling—these tools still represent simplified versions of underlying code-based solutions. To advance the field's use of embeddings, social work education may need to incorporate more technical training, particularly in programming and data science fundamentals.

However, technical skills alone are insufficient. Social work professionals need a nuanced understanding of AI technologies to evaluate and implement embedding-based solutions effectively. This understanding is particularly important given the current discourse around AI risks and limitations. For instance, concerns about AI hallucinations - where models generate false or misleading information - are legitimate for many AI applications, but they don't apply to word embeddings. Embeddings are deterministic mathematical transformations of text; they don't generate new content but rather create fixed numerical representations of existing text. This distinction is crucial for social workers to understand as they evaluate the appropriateness of different AI technologies (see Victor, Goldkind, & Perron, 2024; Rodriguez, Goldkind, L., Victor, Hiltz & Perron, 2024; Singer, Báez, & Rios, 2023).

Additionally, advancing embedding applications in social work requires careful attention to domain-specific needs and challenges. This includes developing specialized embedding models trained on social work literature and documentation, creating validation procedures that align with social work's ethical principles, and establishing best practices for embedding-based analysis in research and practice settings. The field needs frameworks for evaluating embedding models' performance in social work contexts, particularly regarding their ability to accurately capture professional terminology and concepts.

Finally, advancing embeddings in social work requires building bridges between technical experts and social work practitioners. This collaboration ensures that embedding applications are technically sound, practically useful, and ethically implemented. It involves creating channels for knowledge exchange, developing shared vocabularies for discussing these technologies, and establishing feedback loops that allow practitioners to inform the development of embedding-based tools.

## Conclusion

Word embeddings represent a transformative technology for social work research and practice, offering sophisticated approaches to document analysis, information retrieval, and pattern recognition that surpass traditional keyword-based methods. By converting text into mathematical representations preserving semantic relationships, embeddings enable powerful applications from semantic search and cluster analysis to retrieval augmented generation, enhancing research workflows and administrative processes. However, their successful implementation requires careful consideration of inherent limitations, including information loss, training data constraints, interpretability challenges, and potential biases in representation.

Advancing the use of embeddings in social work depends on several critical developments: the creation of domain-specific models trained on social work literature, the development of accessible tools that don't require extensive technical expertise, and the establishment of rigorous best practices for embedding-based analysis that align with social work's ethical principles. Success in these areas requires sustained collaboration between practitioners, researchers, and technology experts to ensure embedding technologies serve the field's unique needs while upholding its core values. As social work evolves in an increasingly data-driven environment, thoughtful integration of embedding technologies can enhance our ability to understand complex patterns in text data, ultimately supporting more effective services and interventions for the individuals and communities we serve.

# References

Association of Social Work Boards. (2024). Exam. Retrieved November 7, 2024, from https://www.aswb.org/exam/

Chase, H., & LangChain Contributors. (2024). LangChain: Building applications with LLMs through composability [Computer software]. https://github.com/langchain-ai/langchain

Chen, J., Xiao, S., Zhang, P., Luo, K., Lian, D., & Liu, Z. (2024). BGE M3-Embedding: Multi-Lingual, Multi-Functionality, Multi-Granularity Text Embeddings Through Self-Knowledge Distillation. arXiv preprint arXiv:2402.03216

Cohen, J. (1992). A power primer. *Psychological Bulletin*, 112(1), 155–159. https://doi.org/10.1037/0033-2909.112.1.155

Cohen, A. M., & Hersh, W. R. (2005). A survey of current work in biomedical text mining. Briefings in Bioinformatics, 6(1), 57-71. https://doi.org/10.1093/bib/6.1.57

Douze, M., Guzhva, A., Deng, C., Johnson, J., Szilvasy, G., Mazaré, P. E., Lomeli, M., Hosseini, L., & Jégou, H. (2024). The Faiss library. arXiv preprint arXiv:2401.0828

Google Cloud. (2024). Ranking documents with Vertex AI Agent Builder. Retrieved November 7, 2024, from https://cloud.google.com/generative-ai-app-builder/docs/ranking

Gu, Y., Tinn, R., Cheng, H., Lucas, M., Usuyama, N., Liu, X., Naumann, T., Gao, J., & Poon, H. (2020). Domain-specific language model pretraining for biomedical natural language processing. arXiv:2007.15779.

Henry, C., Carnochan, S., & Austin, M. J. (2014). Using qualitative data-mining for practice research in child welfare. *Child Welfare*, *93*(6), 7–26. https://doi.org/10.1093/oso/9780197518335.003.0010

Jina AI. (2024). jinaai/jina-embeddings-v3 [Model]. Hugging Face. https://huggingface.co/jinaai/jina-embeddings-v3

Kaushal, A., Mandal, A., Khanna, D., & Acharjee, A. (2023). Analysis of the opinions of individuals on the COVID-19 vaccination on social media. *Digital Health*, *9*, 20552076231186246.

Li, Y., Wehbe, R. M., Ahmad, F. S., Wang, H., & Luo, Y. (2023). A comparative study of pretrained

language models for long clinical text. Journal of the American Medical Informatics Association, 30(2), 340-347.

Liu, J., & LlamaIndex Contributors. (2024). LlamaIndex: A data framework for LLM applications [Computer software]. https://github.com/run-llama/llama_index

Luan, H., Perron, B. E., Victor, B. G., Wan, G., Niu, Y., & Xiao, X. (2024). Using artificial intelligence to support scientometric analysis of scholarly literature: A case example of research on mainland China's left-behind children. *Journal of the Society for Social Work and Research*. Advanced online publication. https://doi.org/10.1086/731613

Mikolov, T., Sutskever, I., Chen, K., Corrado, G., & Dean, J. (2013). Distributed representations of words and phrases and their compositionality. Advances in Neural Information Processing Systems, 26, 3111-3119.

National Social Worker Professional Exam Question Compilation Group. (2024). *Social work comprehensive skills (intermediate) exam questions and answers* (2nd ed.). China Social Sciences Press.

OpenAI. (2024). Embeddings: Learn how to turn text into numbers. Retrieved November 7, 2024, from https://platform.openai.com/docs/guides/embeddings/embedding-models

Perron, B. E., Hiltz, B. S., Khang, E. M., & Savas, S. A. (2024a). AI-enhanced social work: Developing and evaluating retrieval augmented generation (RAG) support systems. *Journal of Social Work Education*. NEED DOI

Perron, B. E., Victor, B. G., Bushman, G., Moore, A., Ryan, J. P., Lu, A. J., & Piellusch, E. K. (2019). Detecting substance-related problems in narrative investigation summaries of child abuse and neglect using text mining and machine learning. *Child Abuse & Neglect*, *98*, 104180. https://doi.org/10.1016/j.chiabu.2019.104180

Perron, B. E., Luan, H., Victor, B. G., Hiltz-Perron, O., & Ryan, J. (2024). Moving Beyond ChatGPT: Local Large Language Models (LLMs) and the Secure Analysis of Confidential

Unstructured Text Data in Social Work Research. *Research on Social Work Practice*, 0(0). https://doi.org/10.1177/10497315241280686

Perron, B. E., Victor, B. G., Hiltz, B. S., & Ryan, J. (2022). Teaching note—data science in the MSW curriculum: Innovating training in statistics and research methods. *Journal of Social Work Education*, *58*(1), 193–198. https://doi.org/10.1080/10437797.2020.1764891

Perron, B. E., Victor, B. G., Ryan, J. P., Piellusch, E. K., & Sokol, R. L. (2022). A text-based approach to measuring opioid-related risk among families involved in the child welfare system. *Child Abuse & Neglect*, *131*, 105688. https://doi.org/10.1016/j.chiabu.2022.105688

Python Software Foundation. (2023). *Python (Version 3.10.9)* [Computer software]. https://www.python.org/

van der Maaten, L., & Hinton, G. (2008). Visualizing data using t-SNE. Journal of Machine Learning Research, 9, 2579-2605.

Victor, B., Goldkind, L. & Perron, B. (2024). Forum: The Limitations of Large Language Models and Emerging Correctives to Support Social Work Scholarship: Selecting the Right Tool for the Task. *International Journal of Social Work Values and Ethics*, 21(1), 200-207. https://doi.org/10.55521/10-021-112

Rodriguez, M. Y., Goldkind, L., Victor, B. G., Hiltz, B., & Perron, B. E. (2024). Introducing generative artificial intelligence into the MSW curriculum: A proposal for the 2029 Educational Policy and Accreditation Standards. *Journal of Social Work Education*, *(60)*2, 174-182. https://doi.org/10.1080/10437797.2024.2340931

Singer, J. B., Báez, J. C., & Rios, J. A. (2023). AI creates the message: Integrating AI language learning models into social work education and practice. *Journal of Social Work Education*, *59*(2), 294-302.

Sokol, R. L., Victor, B. G., Piellusch, E. K., Nielsen, S. B., Ryan, J. P., & Perron, B. E. (2020). Prevalence and context of firearms-related problems in child protective service investigations. *Child Abuse & Neglect*, *107*, 104572. https://doi.org/10.1016/j.chiabu.2020.104572

34Victor, B. G., Kubiak, S., Angell, B., & Perron, B. E. (2023). Time to move beyond the ASWB licensing exams: Can generative artificial intelligence offer a way forward for social work? *Research on Social Work Practice*, *33*(5), 511–517. https://doi.org/10.1177/10497315231166125

Victor, B.G., Goldkind, L., & Perron, B.E. (2024). Forum: The limitations of large language models and emerging correctives to support social work scholarship: Selecting the right tool for the task. *International Journal of Social Work Values and Ethics, 21*(1), 200-207. https://doi.org/10.55521/10-021-112

Victor, B. G., McNally, K., Qi, Z., & Perron, B. E. (2024). Construct-irrelevant variance on the ASWB clinical social work licensing exam: A replication of prior validity concerns. *Research on Social Work Practice*, *34*(2), 217–221. https://doi.org/10.1177/10497315231188305

Victor, B. G., Perron, B. E., Sokol, R. L., Fedina, L., & Ryan, J. P. (2021). Automated identification of domestic violence in written child welfare records: Leveraging text mining and machine learning to enhance social work research and evaluation. *Journal of the Society for Social Work and Research, 12*(4), 631 - 655. https://doi.org/10.1086/712734

**Figure 1. t-Distributed Stochastic Neighbor Embedding (t-SNE) Visualization of Exam Question Embeddings by Test Type**

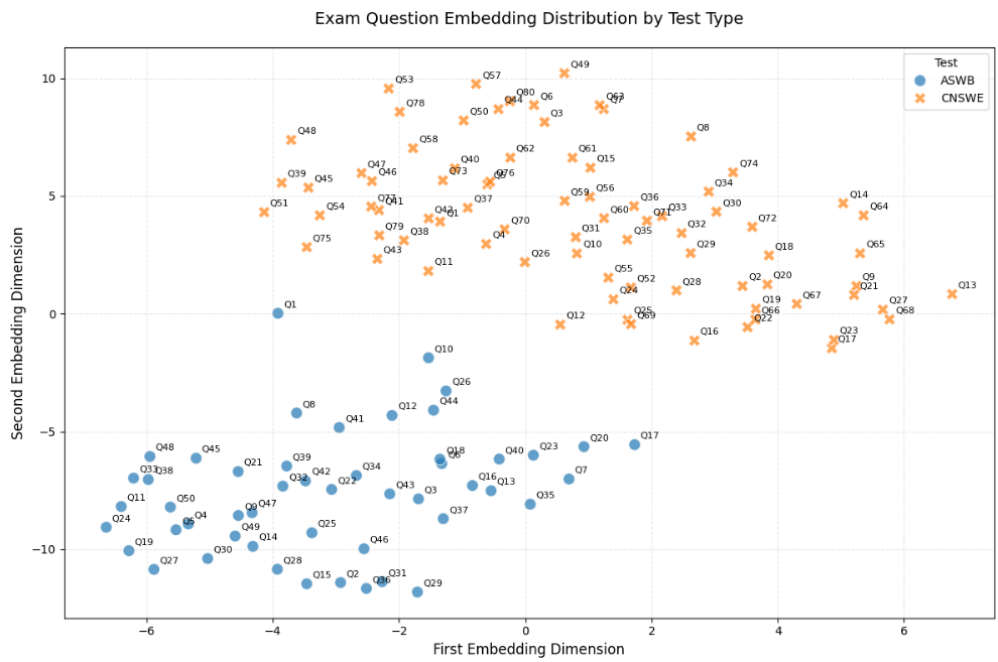

Note: ASWB = Association of Social Work Boards, CNSWE = Chinese National Social Work Examination

**Figure 2: Screenshot of an application to search a database of faculty biographies using vector-based semantic similarity search**

**Figure 3. Example output from an application using vector-based semantic similarity search**

> **Anao Zhang - University of Michigan**
>
> **Name:** Anao Zhang
>
> **Institute:** University of Michigan
>
> **Rank:** Associate Professor
>
> **School / Department:** School of Social Work
>
> **Similarity:** 62.43%
>
> **TLDR-Content:**
> The faculty member's research focuses on psycho-oncology, adolescent and young adult cancer survivorship, and the social determinants of youth health and mental health. They develop and deliver integrated, empirically-supported mental health treatments for individuals with co-occurring physical and mental health conditions.
>
> **Bio:**
> Dr. Anao Zhang is an associate professor of social work at the University of Michigan and the clinical research director of the Adolescents and Young Adults (AYA) Oncology Program at Michigan Medicine. Zhang is a health and mental health intervention researcher with a primary research interest in